\documentclass[conference]{IEEEtran}
\IEEEoverridecommandlockouts
\usepackage{caption}
\usepackage{cite}
\usepackage{amsmath,amssymb,amsfonts}
\usepackage{algorithmic}
\usepackage{graphicx}
\usepackage{textcomp}
\usepackage{xcolor}
\usepackage{multirow}
\def\BibTeX{{\rm B\kern-.05em{\sc i\kern-.025em b}\kern-.08em
    T\kern-.1667em\lower.7ex\hbox{E}\kern-.125emX}}
\begin{document}

\title{Analysis of Robocode Robot Adaptive Confrontation Based on Zero-Sum Game*\\
\thanks{National Natural Science Foundation of China.No.~62076028}
}

\author{\IEEEauthorblockN{\textsuperscript{st} Xiangri LU}
\IEEEauthorblockA{\textit{Automation college} \\
\textit{Beijing University of Technology}\\
Beijing, China \\}
}

\maketitle

\begin{abstract}
The confrontation of modern intelligences is to some extent a non-complete information confrontation, where neither side has access to sufficient information to detect the deployment status of the adversary, and then it is necessary for the intelligences to complete information retrieval adaptively and develop confrontation strategies in the confrontation environment. In this paper, seven tank robots, including TestRobot, are organized for $1V1$ independent and mixed confrontations. The main objective of this paper is to verify the effectiveness of TestRobot's Zero-sum Game Alpha-Beta pruning algorithm combined with the estimation of the opponent's next moment motion position under the game round strategy and the effect of releasing the intelligent body's own bullets in advance to hit the opponent. Finally, based on the results of the confrontation experiments, the natural property differences of the tank intelligences are expressed by plotting histograms of 1V1 independent confrontations and radar plots of mixed confrontations.
\end{abstract}


\section{Introduction:Research background}
Zero-sum game is a concept of game theory, which is a non-cooperative game\cite{b1}. It means that the gain of one party must mean the loss of the other party, and the sum of gain and loss of each party is always "zero", and there is no possibility of cooperation between the two parties \cite{b2,b3,b4,b5}. The result of a zero-sum game is that what one side gains is exactly what the other side loses, and the benefit to society as a whole does not increase by a single point.

The zero-sum game is now an analogue in the sense of strict competition in human society. Nowadays, the two sides of social competition are abstractly represented as the process of zero-sum game. Nowadays, the social competition is complex and changeable, and the two sides confrontation relies more and more on artificial intelligence algorithms to analyze the competitive situation and thus form objective judgments. Machine gaming is an important research direction in the field of artificial intelligence\cite{b6}.In 1997, Deep Blue defeated Kasparov, the king of chess at that time, by one point of total score, after which computers gradually ruled most of chess games except Go\cite{b7,b8,b9}, Deep Blue only has 12 levels of search depth. The basic principle of Deep Blue is similar to the method introduced in this paper, but since the computing speed of Deep Blue is much higher than our current personal computer\cite{b10}. 2016 Google's Alpha Go victory over Korean Go master Lee Sedol made the human-computer game a household name, and then the upgraded Alpha Go Zero defeated the world's top Go player Ke Jie by 4:0 and other events The next upgraded version of Alpha Go Zero with 4:0 victory over the world's top Go player Ke Jie and other events reveal the importance of human-computer gaming. Machine games can be divided into complete information and non-complete information games according to the transparency of information in the game process\cite{b11}. The information of both sides of the complete information game is completely transparent, and the two sides of the game are fully aware of each other's game information. The background of this paper is the hypothesis that both sides of the zero-sum game have game inducing and fraudulent behaviors due to the opaqueness of the information of the non-complete information game, i.e., the game confrontation between the generator and the discriminator in the adversarial neural network\cite{b12,b13}.

Then, in this paper, we simulate the tank robot adversarial environment by Robocode adversarial platform and make the tank robot with Zero-sum Game against the sample robot in Robocode platform, and observe the adversarial data to analyze the adversarial effect produced by Zero-sum Game thinking.
\section{Technical support}

\subsection{Software platform support}

Robocode is a tank robot combat simulation engine released in July 2001 on IBM's Web alpha Works in the U.S. Robocode is a robot combat simulation engine created by Mat Nelson, an IBM engineer, in the Java language\cite{b14}. Robocode is programmed to allow tanks to move, attack, defend, dodge, and fire, while example robots of varying levels of skill are selected from the Robocode counter platform to fight against each other.

\subsection{Algorithm support}
Minimalized maximal search is the core idea of zero-sum games, and minimalized maximal search includes a special Alpha-Beta pruning algorithm\cite{b15,b16,b17}.Alpha Beta pruning algorithm is a safe pruning strategy, that is, it does not have any negative impact on the Robocode platform and the natural properties of the tank robot.Alpha Beta pruning algorithm is based on the fact that tank robots do not take decisions that are detrimental to themselves. If a node in an adversarial environment is clearly a node that is unfavorable to itself, then it can simply prune that node. the Robocode platform tank robot will choose the maximum node at the MAX level, while the newly coded tank intelligence will choose the minimum node at the MIN level. The choice that is unfavorable to both sides then directly cuts this node, i.e. firstly, at the MAX layer, assuming that the current layer has searched for the maximum value, if it finds that the next layer of the next node (which is the MIN layer) will produce a value smaller than the maximum value; secondly, at the MIN layer, assuming that the current layer has searched for the minimum value, if it finds that the next layer of the next node (which is the MAX layer) will produce a value smaller than the minimum value value that is even larger.

According to the analysis of simulation experiments on Robocode platform, when both sides play Zero-sum Game, the opponent always wants to be close to one side to strike so that it can strike the opponent most effectively. Then the process of both tanks playing against each other is represented in the alpha-beta pruning algorithm below\cite{b18}. First of all, both sides confront the environment as shown in the figure. The objective of both sides of the game is to destroy the opponent, and both sides design the idea to move a very small part of the distance to the quadrant where the enemy is located to approach the opponent, but not yet allow the opponent to detect the native tank. Then it is necessary to construct the game circumference and find the optimal path value according to the alpha-beta pruning algorithm of the zero-sum game\cite{b19}. The attacking posture of the attacker is countered.

\begin{figure}[htbp]
\centerline{\includegraphics[width=0.3\textwidth]{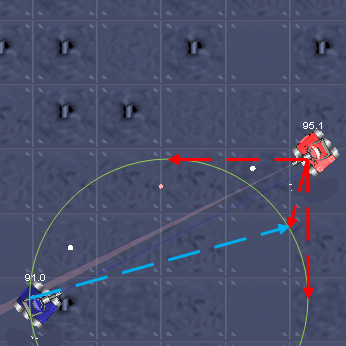}}
\caption{Game confrontation environment.}
\label{fig}
\end{figure}

If the red tank robot wants to attack the blue robot, it must first advance to the white area within a very small area in the figure, then the path to reach the white area is assumed to have N paths in the figure and their curvature varies, and the resulting radius are 20,30,40,70,80,100,120,140,160.

Then for the situation in the figure at this time, the $\alpha-\beta$ pruning algorithm is used to analyze the following process: where the red block represents the red tank maximum Max selection game, and the blue block represents the blue tank minimum Min selection game, then
\begin{figure}[htbp]
\centerline{\includegraphics[width=0.5\textwidth]{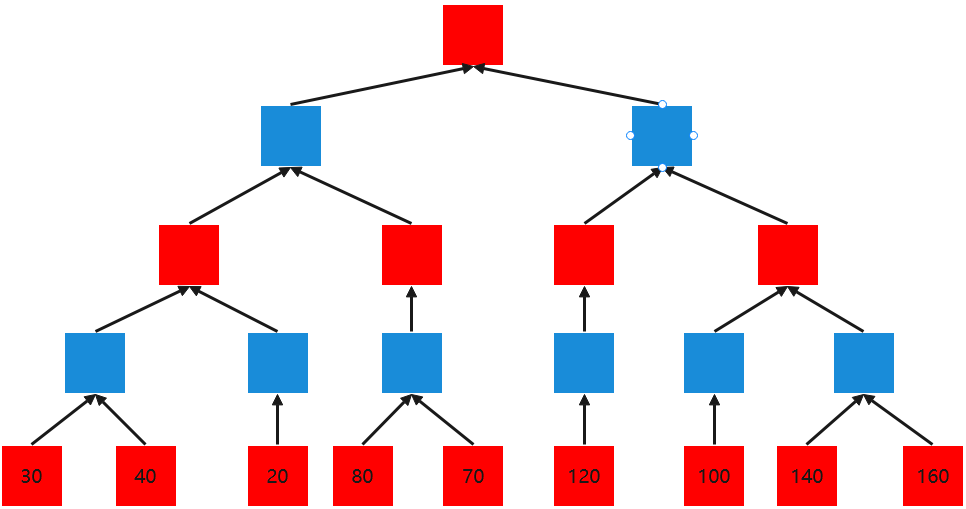}}
\caption{$\alpha-\beta$ pruning algorithm based game tree structure.}
\label{fig}
\end{figure}

If it is known that both tanks are confronted in a region of the map and the backpropagation values of all subnodes of the possible path radius of the confrontation between the two sides can be pushed, the red block represents the red tank maximum Max selection game and the blue block represents the blue tank minimum Min selection game.

If some of the possible path radius sub-nodes of a confrontation region are known, although the backpropagation value of the node cannot be calculated, the range of values of the backpropagation value of the node can be calculated. At the same time, using the range of the backpropagation value of the node, it is not necessary to search for the remaining subnodes if it has been determined that there is no better path when searching for its subnodes. That is, the redundant child nodes are partially cut off.

Specify the positive direction from the bottom to the top, respectively, for 1-5 times the game, as shown in the figure, six nodes in the second layer, four nodes in the third layer, and two nodes in the second layer. Let V be the backward value of the node and $\alpha<V<\beta$, that is, $\alpha$ is the maximum lower bound and $\beta$ is the minimum upper bound. When $\alpha>=\beta$, the remaining branches of the node need not continue searching.

\begin{figure}[htbp]
\centerline{\includegraphics[width=0.5\textwidth]{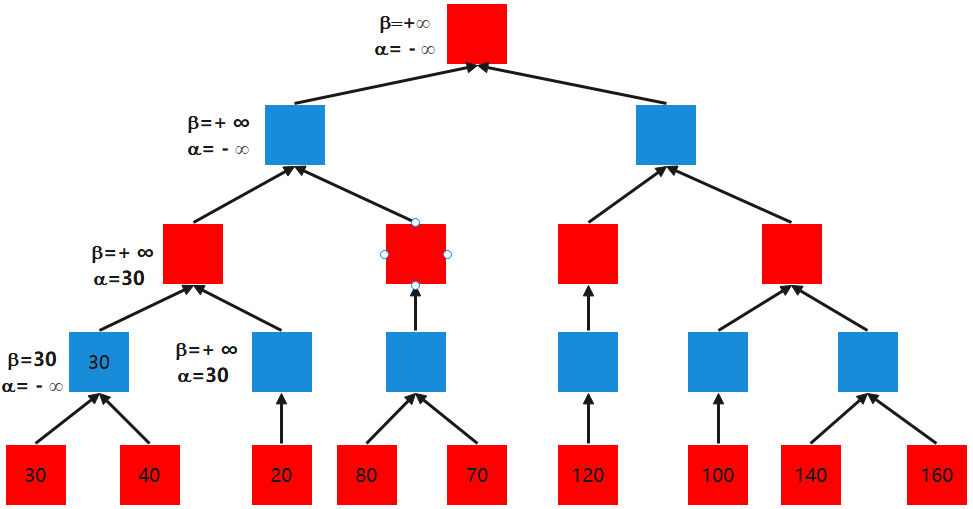}}
\caption{$\alpha-\beta$ pruning game based on the first starting point of the second level.}
\label{fig}
\end{figure}

First, observe the first game extrapolation between the two sides before the confrontation. Initialization, so that $\alpha=-\infty, \beta=+\infty$, that is, $-\infty<V<+\infty$, to the second layer of the first node, due to the left child node of the backpropagation value of 30, and the node is MIN node, try to find the backpropagation value of the small path, so the $\beta$ value is modified to 30, this is because 30 is less than the current $\beta$ value. Then the backpropagation value of the right child of the node is 40, and the value of 30 of the node is not modified at this time, because 40 is larger than the current $\beta$ value. After all the children of the node are searched, the backpropagation value of the node is calculated as 30.

\begin{figure}[htbp]
\centerline{\includegraphics[width=0.5\textwidth]{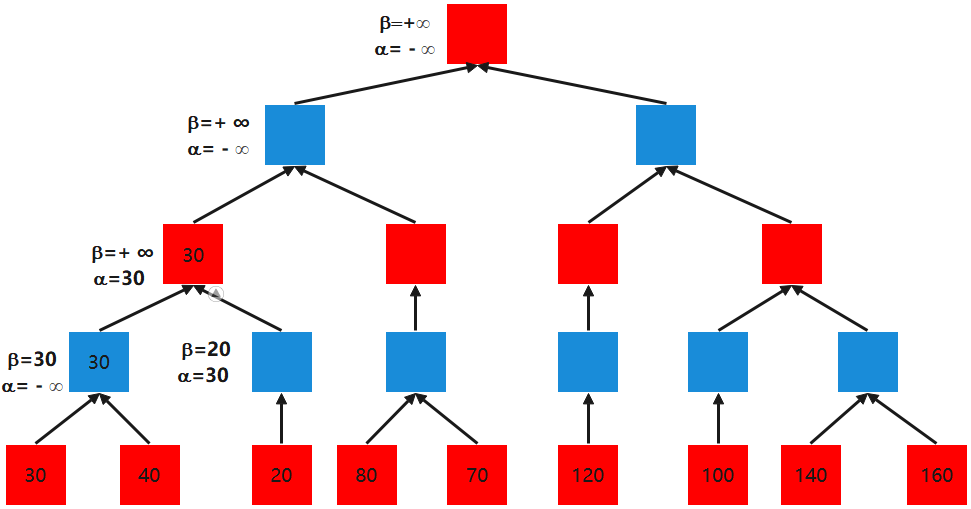}}
\caption{$\alpha-\beta$ pruning game based on the second node of the second layer.}
\label{fig}
\end{figure}

The first node of the second layer is the child of the first node of the third layer, and after calculating the backpropagation value of the first node of the second layer, we can update the backpropagation value range of the first node of the third layer. Since the first node of the third layer is the Max node, we try to find the path with large backpropagation value, so we modify the $\alpha$ value to 30, because 30 is larger than the current $\alpha$ value. After that, the right child node of the first node in the third layer is searched and the $\alpha$ and $\beta$ values of the first node in the third layer are passed to the right child node of the first node in the third layer.

\begin{figure}[htbp]
\centerline{\includegraphics[width=0.5\textwidth]{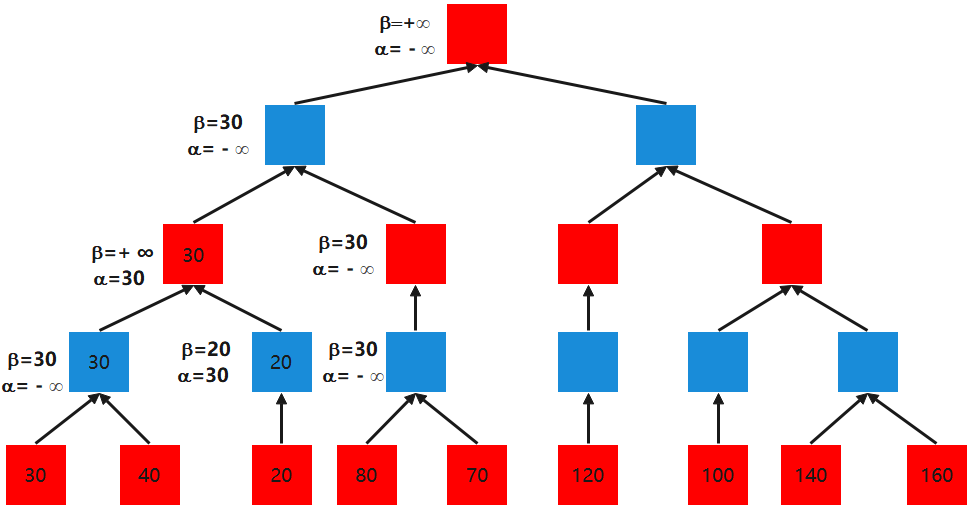}}
\caption{$\alpha-\beta$ pruning game based on the first node of the three layers.}
\label{fig}
\end{figure}

For the second node of the second layer, since there is only one child node, and this node is a Min node, to find the minimum value, the range value of the second node of the second layer is changed to $\alpha$=30 and $\beta$=20, this node violates the pruning algorithm logic, so the right nodes of this node will be cut off. However, the second node in the second layer in the figure does not have a right node, so this step of the operation execution procedure can be omitted. So it can be introduced that the first node path selection in the third layer is 30, and because the second layer is looking for Min path point, then the $\beta$ value will be changed to 30 and passed to the right child node.

\begin{figure}[htbp]
\centerline{\includegraphics[width=0.5\textwidth]{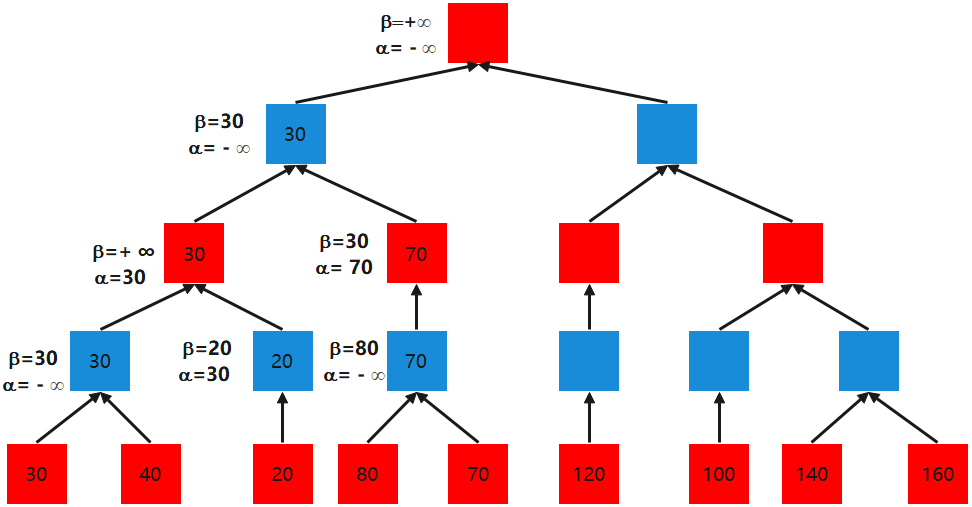}}
\caption{$\alpha-\beta$ pruning game based on the third starting point of the second level.}
\label{fig}
\end{figure}

The third node of the second layer belongs to the search for Min value, then the third node of the second layer has a range value of $\alpha$=$-\infty$ and $\beta$=80, but it can be seen from the figure that the right child node value of the fifth node of the first layer is in the range of $-\infty<\mathrm{V}<80$, then the third node of the second layer takes the value of 70. the second node of the second layer belongs to the search for Max value, then the second node of the third layer has a range value of $\alpha$=70 and $\beta$=30, then the first node in the fourth layer is finally determined to be 30.

\begin{figure}[htbp]
\centerline{\includegraphics[width=0.5\textwidth]{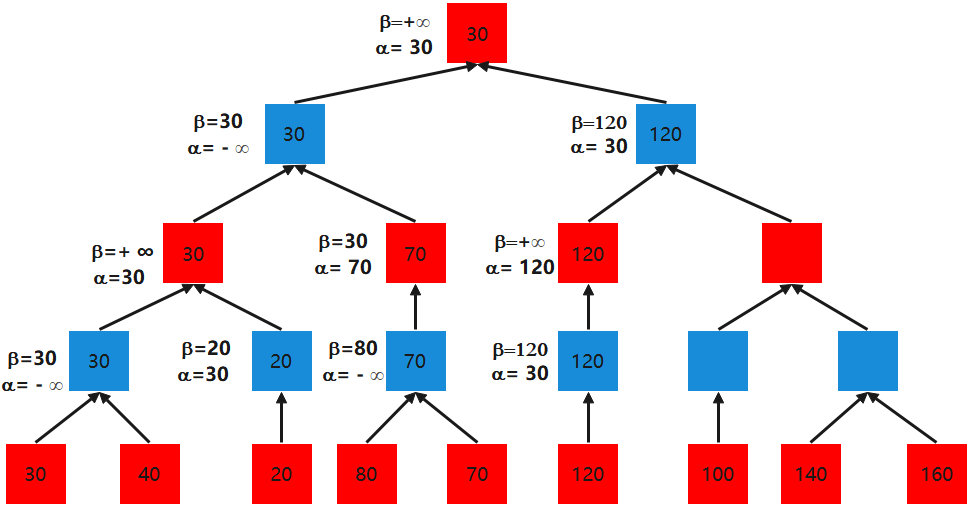}}
\caption{$\alpha-\beta$ pruning game based on the fourth starting point of the second level.}
\label{fig}
\end{figure}

The top node belongs to the search for Max value, then the range value of the top node is $\alpha$=30 and $\beta$=$+\infty$. According to the right node passing law of pruning algorithm, the fifth node of the second layer has a range of $30<\mathrm{V}<+\infty$, while the fifth node of the second layer is replaced with $\beta$ value of 120 according to the passing of the sixth node of the first layer, and the second node of the fourth layer gets the range value of $\alpha$=30 and $\beta$=120 similarly. according to the two nodes of the fourth layer, we can judge that the final value of the top node is 30. at this time, the second node of the fourth layer can be completely pruned. For system integrity, the path traversal tree is analyzed as follows.

\begin{figure}[htbp]
\centerline{\includegraphics[width=0.5\textwidth]{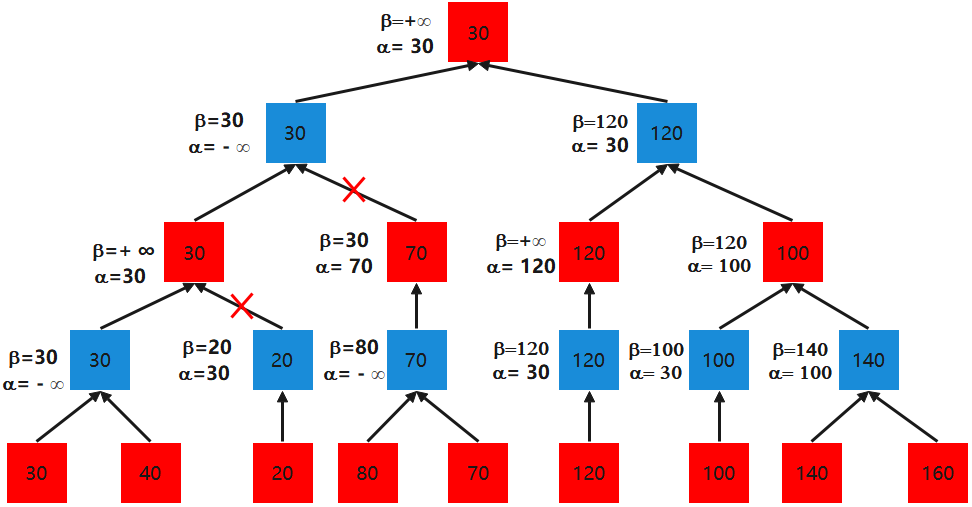}}
\caption{Complete game tree structure based on $\alpha-\beta$ pruning.}
\label{fig}
\end{figure}

The fourth node of the third layer belongs to the search for Max value, then inherit the second node of the fourth layer range value $\alpha$=30, $\beta$=120, similarly the fifth node of the second layer range value inherit $\alpha$=30, $\beta$=120, because this point belongs to the search for Min value, then $\beta$ value is 100, range value $\alpha$=30, $\beta$=100; the fourth node of the third layer belongs to the search for Max value, then $\alpha$ value is 100 The sixth node of the second layer inherits the fourth node of the third layer range value $\alpha$=100, $\beta$=120, the sixth node of the second layer belongs to the search for Min value, then $\beta$=140, according to the pruning rule to select the sixth node of the second layer path value of 140, the fourth node of the third layer path value of 100, the second node of the fourth layer path value is 120.

Through the $\alpha$-$\beta$ pruning algorithm of the Zero-sum Game, the radius of the arc trajectory that the opponent tank chooses to move to a certain area can be initially determined, i.e., the opponent tank robot will most likely choose an arc trajectory with a radius of 30. If corresponding measures are taken, the trajectory of the adversary tank has to be calculated and the strike route of the unmanned tank is designed in advance. For the game problem of nonlinear complex system, a Zero-sum Game against circular algorithm is specially designed.

\begin{figure}[htbp]
\centerline{\includegraphics[width=0.4\textwidth]{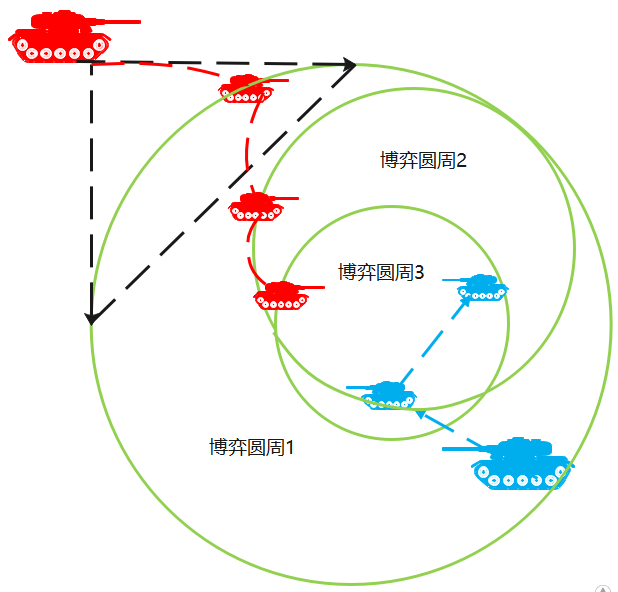}}
\caption{$\alpha-\beta$ pruning based game against rounding.}
\label{fig}
\end{figure}

In summary, the process of $\alpha$-$\beta$ pruning algorithm is only a game selection of the above game circumference 1, game circumference 2 and game circumference 3 still need to repeat the above process of $\alpha$-$\beta$ pruning algorithm, it should be noted that in each game circumference of $\alpha$-$\beta$ pruning algorithm of the first layer of path value selection The upper layer node process is a dynamic change process, then there will be a variety of a game circumference multiple path values of the situation.

\section{Zero-sum Game against rounding algorithm}
As opposed to the attacking side, the defending side needs to construct the corresponding mathematical model for defense, then the construction rules need to be followed as follows.
\begin{itemize}
\item Determine the body of the tank robot, the gun and radar and scanning arc need to be marked with color.
\item Determine the separation of the tank robot's vehicle gun and radar, so that they do not affect each other.
\item Determine the algorithm of tank robot movement.
\item Determine the algorithm of the tank robot to select the firepower.
\item Determine the algorithm for the tank robot to lock on to other robots.
\item Determine the algorithm for the tank robot to calculate and adjust the muzzle of the gun to the enemy.
\end{itemize}

$(1)$ and $(2)$ belong to the basic rules, which can be defined by calling statements inside the Robocode platform, and $(3)$, $(4)$, $(5)$ and $(6)$ belong to the quick calculation design. As shown in the figure

\begin{figure}[htbp]
\centerline{\includegraphics[width=0.4\textwidth]{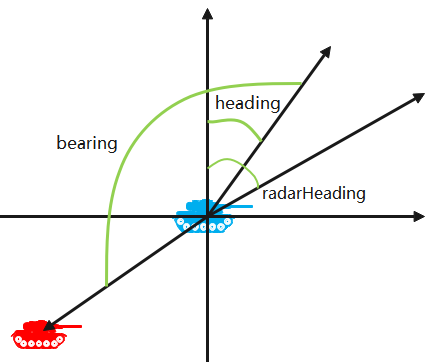}}
\caption{Adversarial coordinate transformation model in games.}
\label{fig}
\end{figure}

The red tank robot is in the third quadrant and the defending robot has to define the angle of the sensor, the positive direction of the tank and the angle between the positive direction of the tank and the positive direction of the red tank in order to build an accurate model.

According to the analysis of the above figure,When the radar scans the enemy, it may get the enemy's pinch angle, and in addition the robot can get its own positive direction angle and the radar's positive direction angle at any time. After the radar has swept the enemy, it will sweep the enemy again to trigger the radar scan event and get the information of the enemy again, and at the same time calculate the angle of the radar back to sweep. Because the enemy's position in the left side of the heading, so the bearing angle is negative, according to the above chart can be analyzed algorithm for $radarHeading$ - $heading$ + $bearing$ can get the robot radar should be back to sweep the angle.

\begin{figure}[htbp]
\centerline{\includegraphics[width=0.4\textwidth]{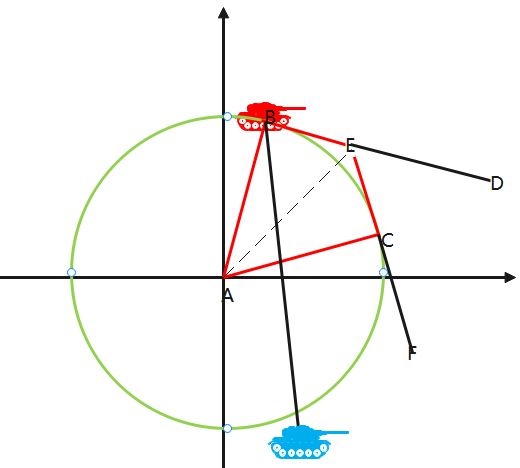}}
\caption{Adversarial circular analytic geometry model of the game process.}
\label{fig}
\end{figure}

If the red tank robot wants to strike the blue tank robot, the length of arc BC is known according to the $\alpha$-$\beta$ pruning algorithm, then the blue side needs to get the radius of circle A if it wants to determine the position of the red tank's movement at the next moment. Then the geometric model is constructed as described below.

\textbf{In the circle $A$ at the center of the circle $A$ as the origin to establish a plane right angle coordinate system, red tank at point $B$, will be moving along the arc $BC$ to point $C$, let $BE$ is the tangent of circle $A$ and extended to $D$, $EC$ is the tangent of circle $A$ and extended to $F$.Connect $AB$ and $AC$ and prove that $\angle \mathrm{BAC}=\angle \mathrm{CED}$.}

\textbf{Proof}

$\because$ both $BE$ and $EC$ are tangents to circle $A$

$\therefore \angle \mathrm{ABE}=\angle \mathrm{ACE}=\frac{\pi}{2}$

$\therefore \angle \mathrm{BAC}+\angle \mathrm{BEC}=\pi$

Also

$\because \angle \mathrm{CED}+\angle \mathrm{BEC}=\pi$

$\therefore \angle \mathrm{BAC}=\angle \mathrm{CED}$

The number of radians of $\angle \mathrm{BAC}$, which is the angle of the tank turning out in the positive direction, is measured by the tank robot sensor.According to the formula $R$ = $arc$ $BC$ / $\angle \mathrm{BAC}$, and according to the tank robot sensor measurement of $AC$ and $X$-axis angle that can be known $C$ point coordinates.

\begin{equation}
X=R * \operatorname{COS} \angle C A X
\end{equation}
\begin{equation}
\mathrm{Y}=\mathrm{R} * \mathrm{SIN} \angle \mathrm{CAX}
\end{equation}

Then the blue tank robot can determine the direction of the bullet in advance for the next moment.

\section{Experimental design and analysis of results}

The experiments in this paper are based on hardware $Intel(R)$ $Core(TM)$ $i5-9400$ $CPU@2.90GHz$ $RAM$ $8GB$, Eclipse IDE for Java Developers and $robocode-1.9.4.3$ tank adversarial platform. Firstly the alpha-beta pruning path algorithm for zero-sum game of red and blue tanks, the natural properties of tanks and the evasive strike algorithm were compiled and constructed using $Eclipse IDE$ for Java Developers. Secondly $robocode-1.9.4.3$ specifies the specifications of the adversarial environment as well as the ammunition rate of fire, etc. The final results are observed using the alpha-beta pruning path algorithm of the zero-sum game against the example robot in $robocode-1.9.4.3$.

\subsection{Independent confrontation}

In the $1V1$ tank battle mode, the maximum number of rounds for each type of tank battle is set to 30, and the tank positions are randomly assigned at the beginning of each round, and the initial tank life value is 100.
In the following, TestRobot tank robots are constructed to confront six groups of typical example robots and observe the results, as shown in Table 1, where the graphs indicate that TestRobot confronts other six types of tank intelligences in five parts: Total Score, Survival, Bullet Damage, Bullet Bonus, and WINS, respectively. The comparative observation of TestRobot tank robot's confrontation status.

Then set the natural properties in $robocode-1.9.4.3$, such as Number of Rounds:30, Gun Cooling Rate:0.1, Inactivity Time:450, Sentry Border Size:100.

\begin{table*}[]
\centering
\caption{Comparison of TestRobot and other Six Types of Tank Intelligences State Confrontation}
\begin{tabular}{ccccccccccccc}
\hline
\multicolumn{3}{c}{Robot Name}                                                                                                                      & \multicolumn{2}{c}{Total Score} & \multicolumn{2}{c}{Survival} & \multicolumn{2}{c}{Bullet Damage} & \multicolumn{2}{c}{Bullet Bonus} & \multicolumn{2}{c}{WINS} \\ \hline
\multicolumn{1}{c|}{\multirow{6}{*}{TestRobot}} & \multicolumn{1}{c|}{\multirow{6}{*}{\begin{tabular}[c]{@{}c@{}}V\\ \\ S\end{tabular}}} & Crazy    & 4573           & 231            & 1400          & 50           & 2388             & 136            & 459             & 2              & 29          & 2          \\
\multicolumn{1}{c|}{}                           & \multicolumn{1}{c|}{}                                                                  & Fire     & 5531           & 380            & 1500          & 0            & 3006             & 376            & 612             & 0              & 30          & 0          \\
\multicolumn{1}{c|}{}                           & \multicolumn{1}{c|}{}                                                                  & My-Robot & 5144           & 326            & 1500          & 0            & 2730             & 317            & 531             & 0              & 30          & 0          \\
\multicolumn{1}{c|}{}                           & \multicolumn{1}{c|}{}                                                                  & V-Robot  & 4192           & 1105           & 1150          & 350          & 2299             & 518            & 412             & 25             & 23          & 7          \\
\multicolumn{1}{c|}{}                           & \multicolumn{1}{c|}{}                                                                  & SpinBot  & 3314           & 1943           & 800           & 700          & 1943             & 933            & 269             & 89             & 16          & 14         \\
\multicolumn{1}{c|}{}                           & \multicolumn{1}{c|}{}                                                                  & Walls    & 5145           & 300            & 1500          & 0            & 2782             & 290            & 557             & 0              & 30          & 0          \\ \hline
\end{tabular}
\end{table*}

\begin{figure}[htbp]
\centerline{\includegraphics[width=0.5\textwidth]{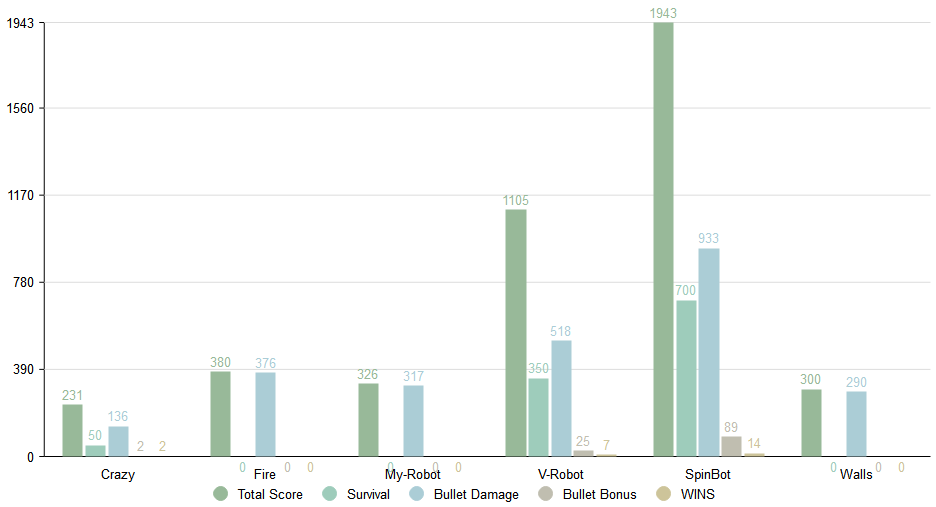}}
\caption{Independent adversarial comparison data between smart bodies.}
\label{fig}
\end{figure}

Among them, it can be observed from the graph that SpinBot can be comparable to TestRobot in terms of strength against Robot tank intelligences, but still cannot beat TestRobot, and the strength of the other four types of tank robots are all far from TestRobot. Through the table, we can calculate the relative strength of TestRobot and Crazy, Fire and other six types of robots, and the relative values of Total Score are $TestRobot$: $Crazy$ = 19.80; $TestRobot$: $Fire$ = 14.56; $TestRobot$: $My-Robot$ = 15.78. $TestRobo$t: $V-Robot$ = 3.79; $TestRobot$: $SpinBot$ = 1.71; $TestRobot$: $Walls$ = 17.15. From the Total Score relative values, it is clear that the TestRobot tank robots have a relative advantage in the adaptive learning environment.

\subsection{Mixed Confrontation}

In order to avoid the chance of experimental results, seven tank counter robots are put into the counter environment at the same time to observe their counter results and analyze them. As shown in Fig.

\begin{figure}[htbp]
\centerline{\includegraphics[width=0.4\textwidth]{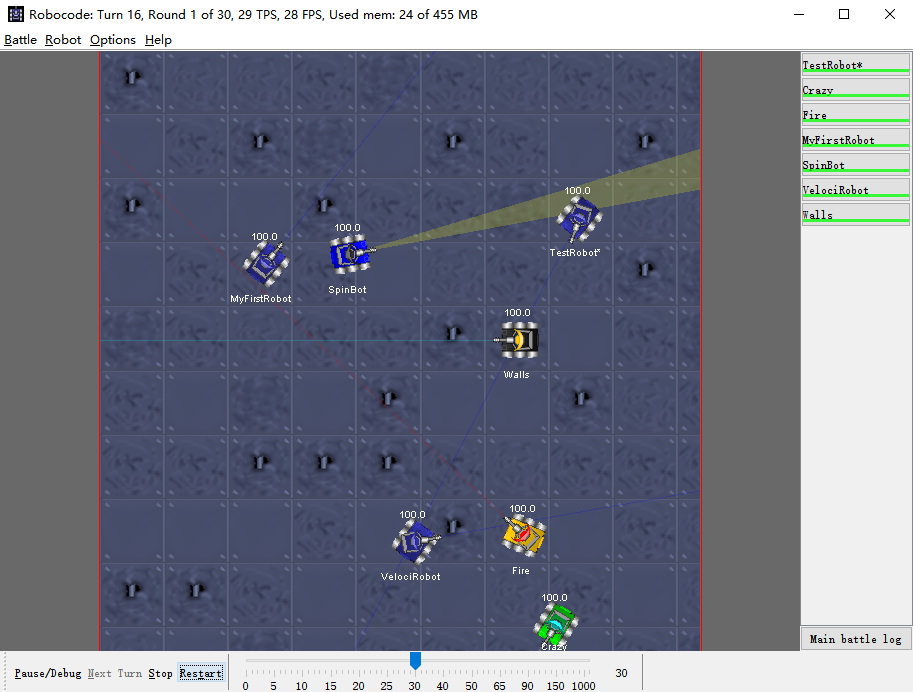}}
\caption{Intelligent body hybrid adversarial gaming environment.}
\label{fig}
\end{figure}

After 30 rounds of confrontation with the seven tank robots in the confrontation environment, the table shows that the $TestRobot$ tank robot can still achieve a good Total Score, the $TestRobot$ and $SpinBot$ tank robots are comparable in terms of survivability, the $Fire$ and $My-Robot$ have the weakest survivability compared to the $TestRobot$, resulting in a lower confrontation score compared to the other tank intelligences. According to the analysis of the Bullet Bonus attribute, $TestRobot$'s hit rate is also very high among the seven tank robots, which leads to a high Bullet Damage score. At the same time, $TestRobot$ was in the top three games 21 times, accounting for 70 percent of the overall.

\begin{figure}[htbp]
\centerline{\includegraphics[width=0.4\textwidth]{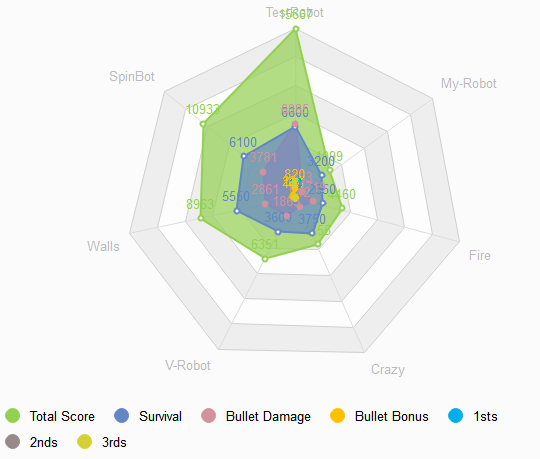}}
\caption{Intelligent body hybrid adversarial data radar map.}
\label{fig}
\end{figure}

\begin{table*}[]
\centering
\caption{Seven types of robots mixed confrontation comparison table}
\begin{tabular}{ccccccccc}
\hline
Rank & Robot Name & Total Score  & Survival & Bullet Damage & Bullet Bonus & 1sts & 2nds & 3rds \\ \hline
1st  & TestRobot  & 15667 (28\%) & 6600     & 6885          & 820          & 14   & 5    & 2    \\
2nd  & SpinBot    & 10933 (20\%) & 6100     & 3781          & 234          & 10   & 2    & 5    \\
3rd  & Walls      & 8963 (16\%)  & 5550     & 2861          & 146          & 5    & 8    & 6    \\
4th  & V-Robot    & 6351 (11\%)  & 3600     & 1863          & 41           & 1    & 6    & 2    \\
5th  & Crazy      & 4855 (9\%)   & 3750     & 942           & 5            & 1    & 1    & 7    \\
6th  & Fire       & 4460 (8\%)   & 2650     & 1719          & 64           & 0    & 2    & 4    \\
7th  & My-Robot   & 4099 (7\%)   & 3200     & 853           & 9            & 0    & 5    & 4    \\ \hline
\end{tabular}
\end{table*}

Combining the above experimental results of independent and mixed confrontation, it can be concluded that $TestRobot$ can overcome the effects of external environmental changes on the intelligent body itself in an adaptive confrontation environment.

\section{Conclusion}
In the unmanned adversarial environment, especially when the adversarial environment is complex and the access to information from both sides is narrow, and the intelligences themselves are required to try to explore the surrounding adversarial environment, the self-adaptation ability of the intelligences based on the zero-sum game adversarial algorithm is verified to be higher than other intelligences through the $Robocode$ tank robot adversarial platform, specifically to design a $TestRobot$ tank intelligences The robot, firstly, uses the $\alpha$-$\beta$ pruning algorithm to select a small range of moving paths, after that, judges the moving direction of the tank robot based on the Zero-sum Game circumference, and finally solves the coordinate points where the opponent may move through the game circumference, making its own intelligent body hit the opponent with advance bullets in advance.

\section*{Acknowledgment}

Project supported by the National Natural Science Foundation of China (No.~62076028)

\clearpage

\begin{appendices}
\textbf{Appendix}

Due to the diversity of path choices in a small range of the attacker's tank intelligence, nine path choices are allowed when the simulation can be achieved and is not general, then there are multiple expressions of the game tree for both sides of the confrontation, here to simplify the model, while starting with the bottom branch on the left side, the left starting game branch on the second level lists three $\alpha$-$\beta$ pruning algorithm cases, which are sequential ( from small to large), inverse order (from large to small), and single-branch maximum starting game.

\begin{figure}[htbp]
\centerline{\includegraphics[width=0.4\textwidth]{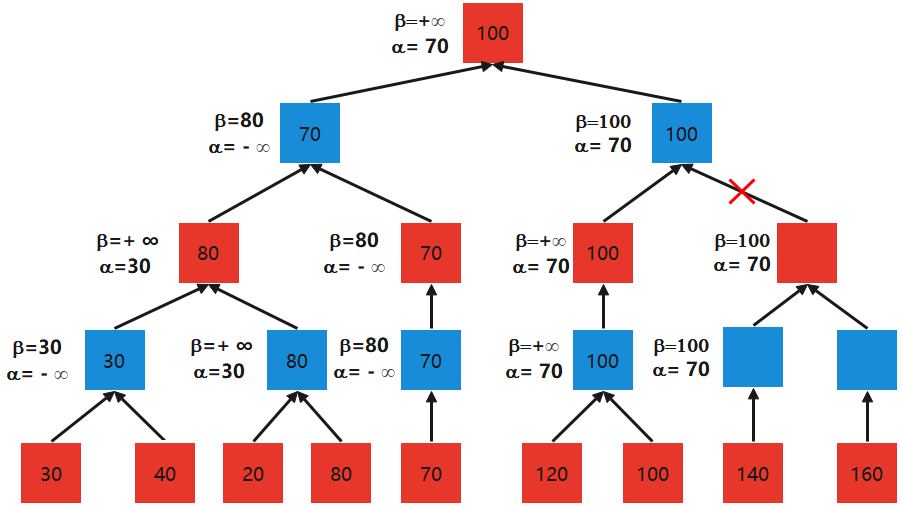}}
\caption{Starting game branching paths from small to large $\alpha$-$\beta$ pruning games.}
\label{fig}
\end{figure}

First of all the second layer of the left side of the starting game branch is the order (from small to large), according to the principle of the two sides of the confrontation to carry out three game rounds constructed the game tree, it can be observed that after screening by $\alpha$-$\beta$ pruning algorithm, the attacker's game path selection range has been clearly marked in the figure, the path value of the second game branch point of the four layers not in the range of the right branch cut off, this game tree final path selection is 100.

\begin{figure}[htbp]
\centerline{\includegraphics[width=0.4\textwidth]{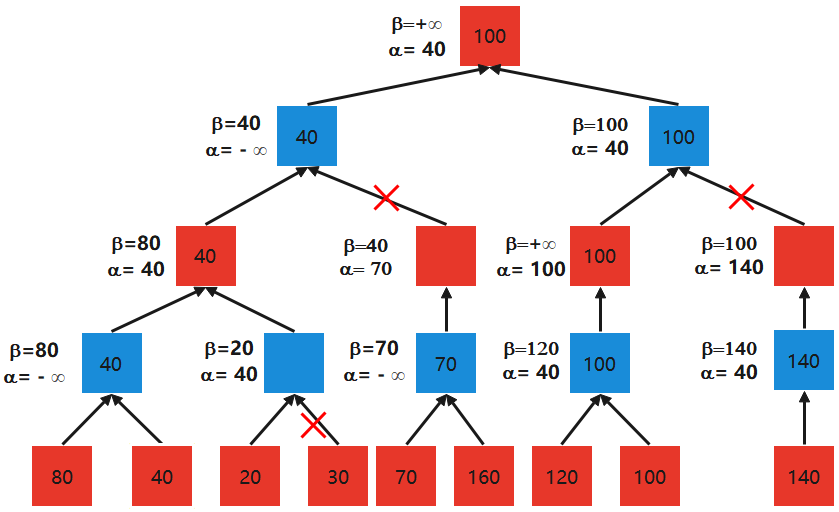}}
\caption{Starting game branching paths from large to small $\alpha$-$\beta$ pruning games.}
\label{fig}
\end{figure}

The left starting game branch of the second layer of this game tree is in reverse order (from largest to smallest), and the game tree is constructed according to the principle that the two sides confront each other for three game rounds, and it can be observed that after screening by $\alpha$-$\beta$ pruning algorithm, the game path selection range of the attacker is clearly marked in the figure, and because of the contradictory path range of the second game branch point of the second layer of this game tree and all branch points of the fourth layer. The final path selection of this game tree is 100 by cutting off their right branches.

\begin{figure}[htbp]
\centerline{\includegraphics[width=0.4\textwidth]{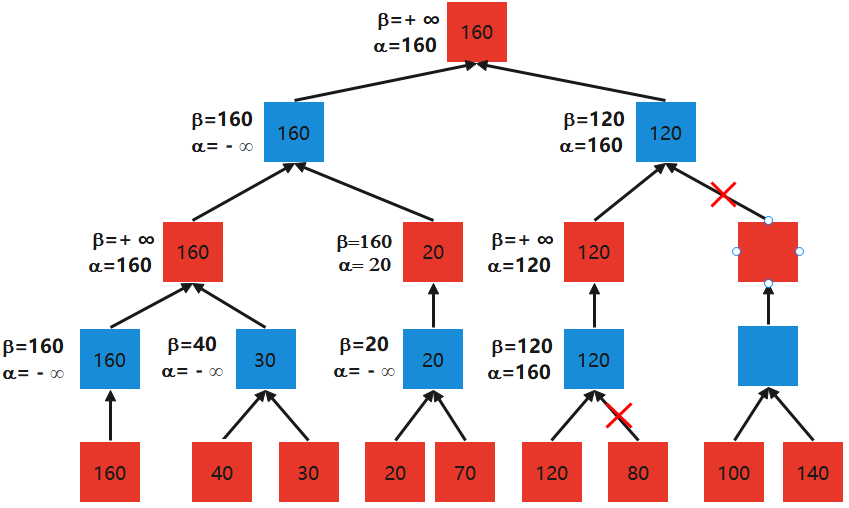}}
\caption{$\alpha$-$\beta$ pruning game with single branch maxima of the starting game branching path.}
\label{fig}
\end{figure}

The left starting game branch of the second layer of this game tree is a single branch maximum, and the game tree is constructed according to the principle that both sides confront each other for three game rounds, and it can be observed that after screening by $\alpha$-$\beta$ pruning algorithm, the game path selection range of the attacker is clearly marked in the figure, and because of the contradictory path range of the fourth game branch point of the second layer of this game tree and the second branch point of the fourth layer, their The final path choice of this game tree is 160.
\end{appendices}


\begin{thebibliography}{00}
\bibitem{b1} Lu, Xiangri, Wang, Zhanqing, Ma, Hongbin. Cost function selection and performance evaluation in zero-sum game adversarial. Microelectronics and Computers,2021,38-07:30-35.DOI:10.19304/j.cnki.issn1000-7180.2021.07.006.
\bibitem{b2} Atsuhiro Satoh,Yasuhito Tanaka. Sion's minimax theorem and Nash equilibrium of symmetric three-players zero-sum game[J]. International Journal of Mathematics in Operational Research,2020,16-2:
\bibitem{b3} Max S Kim. ZERO-sum game[J]. MIT Technology Review,2020,123-1
\bibitem{b4} Atsuhiro Satoh,Yasuhito Tanaka. Two Person Zero-Sum Game with Two Sets of Strategic Variables[J]. International Game Theory Review,2019,21-03:
\bibitem{b5} Bu?mann Peter. The Result is a Zero-Sum Game.[J]. Deutsches Arzteblatt international,2019,116-1-2:
\bibitem{b6} Zhang Z,Pang H. A study of machine games and their search algorithms[J]. Software Guide,2008-07:48-50.
\bibitem{b7} Liu, Jia-Yao, Lin, Tao. Design of black and white chess gaming system[J]. Intelligent Computers and Applications,2020,10-05:176-179+182.
\bibitem{b8}Zhang Xiaomian. Research and design of a computer gaming system for backgammon[D]. Anhui University,2017.
\bibitem{b9}Dong Huiying,Wang Yang. Research on multiple search algorithms for backgammon gaming[J]. Journal of Shenyang University of Technology,2017,36-02:39-43+83.
\bibitem{b10}Putra Werda Buana,Heryawan Lukman. APPLYING ALPHA-BETA ALGORITHM IN A CHESS ENGINE[J]. Jurnal Teknosains,2017,6-1:
\bibitem{b11}O. V. Baskov. Bounded Computational Capacity Equilibrium in Repeated Two-Player Zero-Sum Games[J]. International Game Theory Review,2017,19-3:
\bibitem{b12}Fabien Gensbittel,Christine Grun. Zero-Sum Stopping Games with Asymmetric Information[J]. Mathematics of Operations Research,2019,44-1:
\bibitem{b13}Misha Gavrilovich,Victoria Kreps. Games with Symmetric Incomplete Information and Asymmetric Computational Resources[J]. International Game Theory Review,2018,20-2):
\bibitem{b14}Robocode[EB/OL].https://robowiki.net/wiki/Robocode. 16 September 2017
\bibitem{b15}Wang Zengcai. Design and development of Othello game based on Alpha-Beta pruning algorithm[D]. Inner Mongolia University,2016.
\bibitem{b16}Liu Shuying,Mu Yuanbiao,Li Hong. Design of Chinese chess game based on alpha-beta pruning search algorithm[J]. Information Communication,2015-08:47-48.
\bibitem{b17}Zheng Jianlei,Kuang Fangjun. Research and implementation of intelligent game algorithm based on Minimal Great Value Search and Alpha Beta pruning algorithm for five chess[J]. Journal of Wenzhou University Natural Science Edition),2019,40-03):53-62.
\bibitem{b18}Sylvain Sorin,Guillaume Vigeral. Limit Optimal Trajectories in Zero-Sum Stochastic Games[J]. Dynamic Games and Applications,2019,10 prepublish:
\bibitem{b19}Senthuran Arunthavanathan,Leonardo Goratti,Lorenzo Maggi,Francesco de Pellegrini,Sithamparanathan Kandeepan,Sam Reisenfield. An optimal transmission strategy in zero-sum matrix games under intelligent jamming attacks[J]. Wireless Networks,2019,25-4:
\end{thebibliography}
\end{document}